\documentclass{article}

\usepackage[preprint]{corl_2026} 

\usepackage{amssymb}
\usepackage{amsmath} 
\usepackage{titlesec}
\usepackage{algpseudocode}
\usepackage{graphicx}
\usepackage[square,numbers]{natbib}
\usepackage[utf8]{inputenc} 
\usepackage[T1]{fontenc}    
\usepackage{hyperref}       
\usepackage{url}            
\usepackage{booktabs}       
\usepackage{amsfonts}       
\usepackage{nicefrac}       
\usepackage{microtype}      
\usepackage{booktabs}
\usepackage{adjustbox}
\usepackage{algorithm}
\usepackage{subcaption}
\usepackage{caption}
\usepackage{enumitem}
\usepackage{multirow}
\usepackage{wrapfig}

\usepackage[dvipsnames]{xcolor}

\title{ReGuide: From Test-Time Guidance to \\ Self-Improving Diffusion Policies}

\author{
  Tzu-Hsiang Lin, Srinivas Shakkottai, Dileep Kalathil, and P. R. Kumar \\
  Texas A\&M University \\
  \texttt{\url{https://reguide-project.github.io}}
}

\begin{document}
\maketitle


\begin{abstract}
    Behavior-cloned diffusion policies are expressive but remain vulnerable to covariate shift: small deviations from demonstrated states can compound into task failure. Existing methods address this either by expanding the training distribution through expert corrections or synthetic augmentation, or by steering a frozen policy at test time with guidance from a learned model. The former can be expensive or assumption-dependent, while the latter discards the corrected trajectories after execution. We introduce \textbf{ReGuide}, a self-improving framework that treats guided rollouts as reusable on-policy recovery data. ReGuide first uses \textbf{Phase-Conditioned Guidance (PCG)} to generate corrective rollouts: it constructs phase-specific latent targets, applies guidance only in the drifted-but-recoverable regime, and guides through the estimated clean action to match the dynamics model's training distribution. Successful guided rollouts are then absorbed back into the policy through \textbf{ReGuide-FT}, which fine-tunes the current checkpoint, or \textbf{ReGuide-FS}, which retrains from scratch on the augmented dataset; the two can also be composed and iterated. On Robomimic Can, Square, Transport, and Tool Hang, ReGuide improves base-policy success by $1.3$--$7.7\times$, outperforms LPB in the test-time-only setting, and matched-data ablations show that the gains come from guided recovery data rather than additional rollouts alone.
\end{abstract}



\section{Introduction}

Imitation learning-based policies are vulnerable to covariate shift: small errors move the robot away from demonstrated states, after which the policy is queried on observations outside its training distribution and errors compound over time~\citep{ross2011dagger}. This problem is especially acute for long-horizon manipulation, where an early deviation can make later subtasks unrecoverable. Diffusion policies improve the expressivity by modeling multimodal action distributions~\citep{chi2024diffusionpolicy}, but they do not remove this distribution-shift problem. A common solution is to expand the training distribution. DAgger-style methods collect expert corrections at learner-visited states~\citep{ross2011dagger, hu2025rac, xu2025crdagger}, while data-augmentation methods synthesize additional examples using task structure, simulation, or local dynamics assumptions~\citep{ke2024ccil, ankile2024juicer, jia2022seil}. These methods address the right failure mode, but require either continued expert access or assumptions about which perturbations preserve task correctness. In contrast, recent test-time guidance methods such as DynaGuide~\citep{du2025dynaguide} and Latent Policy Barrier (LPB)~\citep{sun2025lpb} steer a frozen diffusion policy using gradients from a learned dynamics model, improving recovery without retraining or new demonstrations. However, these methods discard the guided trajectory after execution.

In this paper, we propose a simple yet highly effective solution for this problem: \textbf{use guided rollouts as training data for iterative self-improvement}. A successful guided rollout captures precisely the information missing from the original demonstrations: where the learned policy tends to drift, and which corrective actions return the system toward task completion. Recycling such rollouts provides on-policy recovery data using only a trajectory-level success signal, rather than expensive per-state expert relabeling as used in Dagger-style algorithms.

Although this idea is simple to state, turning guided rollouts into useful training data requires several technical choices that prevent self-generated data from degrading the policy. Guided rollouts are useful only when the guidance signal is reliable. Long-horizon tasks are phase-structured, and a single global target set can pull the rollout toward the wrong stage or collapse valid behavior modes. Dynamics gradients are also reliable only near the data distribution: guidance applied too close to the expert manifold can perturb correct actions, while guidance applied too far away can rely on extrapolative predictions. Finally, standard diffusion guidance differentiates through noisy denoising iterates, whereas the dynamics model is typically trained on clean action chunks.

We introduce \textbf{ReGuide}, a self-improving diffusion-policy framework that turns this idea into a practical algorithm by controlling how guided rollouts are generated, filtered, and absorbed back into the policy. ReGuide first uses \textbf{Phase-Conditioned Guidance (PCG)} to generate reliable corrective rollouts: PCG constructs phase-specific latent targets from demonstrations, preserves multimodality through multiple targets per phase, gates guidance to the drifted-but-recoverable regime, and applies the guidance gradient through the estimated clean action following MPGD~\citep{he2023mpgd}. The successful guided rollouts are then recycled into policy training through two complementary update mechanisms: \textbf{ReGuide-FT}, which fine-tunes the current checkpoint with a rehearsal-style mixture of demonstrations and guided rollouts, and \textbf{ReGuide-FS}, which retrains a fresh policy on the augmented dataset. Crucially, the updated policy can be rolled out again under the same guidance mechanism to generate a new batch of higher-quality recovery data, yielding an iterative rollout--collect--train loop that improves the policy without additional expert demonstrations until the self-generated data reaches its performance ceiling.

On Robomimic tasks~\citep{mandlekar2021robomimic}, ReGuide improves base-policy success by $1.3$--$7.7\times$. PCG outperforms LPB in the test-time-only setting, matched-data ablations show that guided rollouts are more valuable than unguided rollouts at the same data volume, and a second ReGuide-FT iteration further improves performance, showing that guided rollout generation can support iterative self-improvement.

\vspace{-0.3cm}
\section{Related Work}
\label{sec:related_work}
\vspace{-0.2cm}
\textbf{Diffusion policies and test-time guidance.}
We build on Diffusion Policy~\citep{chi2024diffusionpolicy} as the base visuomotor policy. Recent work steers pretrained diffusion policies at inference time without modifying their weights: DynaGuide~\citep{du2025dynaguide} uses gradients from an external dynamics model, Latent Policy Barrier (LPB)~\citep{sun2025lpb} treats expert latent embeddings as an implicit in-distribution barrier, PPGuide~\citep{wang2026ppguide} learns a performance predictor for guidance, and Generative Predictive Control (GPC)~\citep{qi2026gpc} performs model-based look-ahead. These methods improve execution-time behavior, but the guided trajectories are discarded after each episode. ReGuide instead uses guidance as a data-generation mechanism: successful guided rollouts are added back to the training set to improve the policy. ReGuide also differs in the guidance design: we adapt Manifold Preserving Guided Diffusion~\citep{he2023mpgd} to differentiate through the estimated clean action, and replace global targets with phase-conditioned target sets and a two-threshold gate. LPB is the closest architectural baseline, since it also applies dynamics-gradient guidance to a diffusion policy in latent space.

\textbf{Self-improvement and data augmentation for imitation learning.}
Covariate shift in behavior cloning is commonly addressed by expanding the learner's training distribution. Interactive methods such as DAgger~\citep{ross2011dagger} and RaC~\citep{hu2025rac} collect new supervision at policy-visited states, but require continued expert access. Other approaches augment the data without further expert intervention~\citep{ke2024ccil, ankile2024juicer, jia2022seil, mehta2025stablebc}. The closest in spirit is CCIL~\citep{ke2024ccil}, which learns a Lipschitz-regularized dynamics model from demonstrations and generates corrective action labels near demonstrated states. ReGuide differs in both mechanism and loop structure: it uses test-time dynamics gradients to recover rollouts that have already drifted from the demonstrations, filters successful trajectories, and iteratively retrains or fine-tunes the policy on the resulting on-policy recovery data.

\textbf{Other related work.} Extended discussion of related work is provided in Appendix~\ref{sec:extended_related_work}.

\section{Preliminaries}
 
\textbf{Diffusion policy.}  Given expert demonstrations $\mathcal{D}_{\text{expert}} = \{(o_t, a_t)_{t=1}^{T}\}$, where $o_t \in \mathcal{O}$ and $a_t \in \mathcal{A}$ denote the observation and action at time $t$, an imitation learning algorithm learns a policy $\pi_\theta : \mathcal{O} \rightarrow \mathcal{A}$ by minimizing a supervised loss. A policy is called a  \emph{diffusion policy}~\citep{chi2024diffusionpolicy} when $\pi_\theta$ is parameterized as a conditional denoising diffusion model, conditioned on the past observations. Following~\citet{chi2024diffusionpolicy}, the policy conditions on a window of $T_o$ past observations $O_t = (o_{t-T_o+1}, \dots, o_t)$ and generates a chunk of $T_p$ future actions $A_t = (a_t, a_{t+1}, \dots, a_{t+T_p-1})$, of which the first few are executed before re-planning. In practice, the denoiser conditions on a low-dimensional latent embedding $z_t = h(O_t)$ rather than the raw observation window.

To sample an action chunk $A_t \sim \pi_\theta(\cdot \mid O_t)$, a diffusion policy starts from Gaussian noise $A_t^K \sim \mathcal{N}(0, I)$ and iteratively denoises through $K$ reverse-diffusion steps. At each step $k$, a learned noise predictor $\epsilon_\theta(A_t^k, k, z_t)$ produces the reverse update
\begin{equation}
\label{eq:reverse-step}
    A_t^{k-1} \;=\; \frac{1}{\sqrt{\alpha_k}}\!\left( A_t^k 
    \;-\; \frac{1 - \alpha_k}{\sqrt{1 - \bar{\alpha}_k}}\, 
    \epsilon_\theta(A_t^k, k, z_t) \right) \;+\; \sigma_k\, \xi,
    \qquad \xi \sim \mathcal{N}(0, I),
\end{equation}
where $\alpha_k$, $\bar{\alpha}_k$, and $\sigma_k$ are determined by the noise schedule and $\xi$ is set to zero at the final step ($k = 1$). Iterating from $k = K$ down to $k = 1$ yields the executed action chunk $A_t = A_t^0$.

\textbf{Guidance for diffusion policies.} Diffusion models admit a natural mechanism for steering generation at inference time: the reverse process can be biased toward a desired property with the gradient of a differentiable reward function. Concretely, when generating a sample $x$ via 
reverse diffusion from $x_K \sim \mathcal{N}(0, I)$ down to $x_0$, 
classifier guidance~\citep{dhariwal2021classifier} and its classifier-free 
variants~\citep{ho2022classifierfree} modify the predicted noise at step 
$k$ as
\begin{equation}
    \hat{\epsilon}(x_k) \;=\; \epsilon_\theta(x_k) \;-\; \eta\sqrt{1-\bar{\alpha}_k}\;
    \nabla_{x_k}\, r(x_k),
\end{equation}
where $r$ is a reward (e.g., a classifier log-probability for a target 
class, text-to-image fidelity for the generated image), and  $\eta$ controls the guidance strength. The gradient $\nabla_{x_k} r(x_k)$ pushes each denoising step 
toward samples that score higher under $r$, so the final $x_0$ tends to 
satisfy the reward criterion while still lying on the data manifold 
learned by $\epsilon_\theta$. This principle has been used to steer image 
generation toward target classes, text descriptions, and aesthetic 
properties without retraining the underlying diffusion model.

Recent work transfers this idea from image generation to control policies 
represented as diffusion models. In offline reinforcement learning, 
several methods guide a diffusion policy using the gradient of a learned 
value function~\citep{frans2025cfgrl, lu2023qgpo, hansenestruch2023idql}, 
steering action generation toward high-value behavior. In imitation 
learning, DynaGuide~\citep{du2025dynaguide} and LPB~\citep{sun2025lpb} 
instead use a learned dynamics model to keep rollouts close to the 
expert distribution. The general form mirrors the image-generation case 
above: at denoising step $k$, the noise prediction is augmented with the 
gradient of a differentiable cost $\delta$,
\begin{equation}
    \hat{\epsilon}(A_t^k) \;=\; \epsilon_\theta(A_t^k) \;-\; \eta\sqrt{1-\bar{\alpha}_k}\;
    \nabla_{A_t^k}\, \delta(A_t^k),
\end{equation}
with $x_k \leftrightarrow A_t^k$ and the reward $r$ replaced by a 
cost $\delta$ that penalizes drift from the expert manifold. 
The intuition is the same as before: each denoising step is nudged 
toward actions whose predicted consequences score well under 
$\delta$.

\textbf{Dynamics model.} A dynamics model predicts how the environment evolves under a given action: given the latent state $z_t$ and a candidate action chunk $A_t$, it
returns the latent state the environment is expected to reach, $   \hat{z}_{t+T_p} = d_\phi(z_t,\; A_t),$, where  $z_t = h(o_{t})$, and   $h$ is the visual encoder shared with the diffusion policy. Following \citep{sun2025lpb}, we use the cost function $\delta(A_t^k) \;=\; \bigl\|\, d_\phi(z_t, A_t^k) - z^*\,\bigr\|^2,$ where $z^*$ is a target latent state. The gradient $\nabla_{A_t^k}\delta$ 
then pushes each denoising step toward action chunks whose predicted 
consequences lie close to $z^*$, shifting the guidance signal from 
``does this action look like an expert action'' to ``does this action 
lead to an expert-like future state.''

We adopt the DINO-WM architecture~\citep{zhou2025dinowm} for $d_\phi$, 
following~\citep{sun2025lpb}. The dynamics model shares its visual encoder $h$ with the diffusion policy so that latent-space distances are meaningful as a 
guidance signal, and is trained on rollouts (both successful and failed) 
generated by the diffusion policy.


\section{Our Approach: ReGuide}
\vspace{-0.2cm}

\begin{wrapfigure}{r}{0.42\textwidth}
\vspace{-1cm}
\begin{minipage}{0.42\textwidth}
\begin{algorithm}[H]
\caption{ReGuide}
\label{alg:reguide}
\begin{algorithmic}[1]
\small
\Require Demonstration data $\mathcal{D}_{\text{demo}}$, number of iterations $N$
\State $\mathcal{D}_1 \gets \mathcal{D}_{\text{demo}}$
\State Train $\pi_1$ on $\mathcal{D}_{\text{1}}$
\State Construct phase targets and distance distributions via Alg.~\ref{alg:target-construction}
\For{$i = 1$ to $N$}
    \State Collect successful rollouts $\mathcal{D}_{i}^g$ from $\pi_i$ via Alg.~\ref{alg:phase-rollout}
    \State $\mathcal{D}_{i+1} \gets \mathcal{D}_i \cup \mathcal{D}_{i}^g$
    \State Train $\pi_{i+1}$ using $\mathcal{D}_{i+1}$ and $\pi_i$
\EndFor
\State \Return $\pi_{N+1}$
\end{algorithmic}
\end{algorithm}
\end{minipage}
\vspace{-0.3cm}
\end{wrapfigure}
ReGuide turns guidance into a data-generation mechanism for a behavior-cloned diffusion policy without collecting additional expert demonstrations. It consists of three coupled components. First, \textit{phase-aware target construction} (Section~\ref{sec:target_construction}) decomposes long-horizon demonstrations into temporally ordered phases and constructs a set of latent targets for each phase, so that ``in-distribution'' behavior is defined locally. Second, \textit{phase-conditioned guidance and data collection} (Section~\ref{sec:guidance}) uses a learned dynamics model to guide the diffusion denoising process toward the appropriate phase targets, but only in regimes where the predicted future latent is drifted from the demonstrations while still within the model's reliable operating range. Third, \textit{iterative self-improvement learning} (Section~\ref{sec:continual}) filters successful guided rollouts, merges them with the existing dataset, and updates the policy through either fine-tuning or retraining. The updated policy then generates a new distribution of guided rollouts, yielding an iterative rollout--collect--train loop driven by self-generated, success-filtered data.

\begin{figure}[t]
    \centering
    \includegraphics[width=\linewidth]{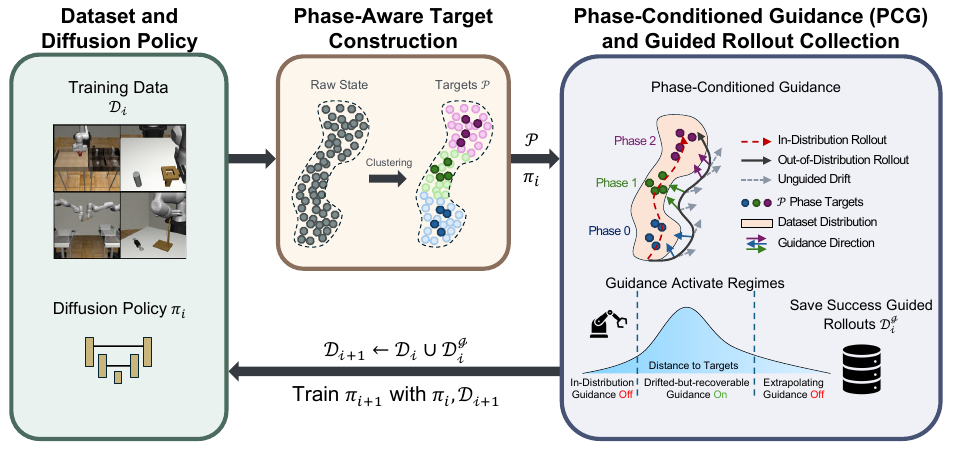}
    \caption{\textbf{ReGuide overview.} At iteration $i$: starting with  $\pi_i$ and $\mathcal{D}_i$, construct phase targets $\mathcal{P}$ by clustering the latent states, roll out $\pi_i$ with phase-conditioned guidance (active only in the drifted-but-recoverable regime per the per-phase distance distribution), and merge successful guided rollouts $\mathcal{D}_i^g$ into $\mathcal{D}_{i+1}$ for the next iteration. Update the policy to get  $\pi_{i+1}$ from $\mathcal{D}_{i+1}$ and  $\pi_{i}$}
    \label{fig:flow}
    \vspace{-0.4cm}
\end{figure}

Algorithm~\ref{alg:reguide} and  Figure~\ref{fig:flow} summarize our ReGuide Algorithm.

\vspace{-0.2cm}
\subsection{Phase-Aware Target Construction}
\label{sec:target_construction}
\vspace{-0.2cm}

\begin{figure}
\centering
\includegraphics[width=\linewidth]{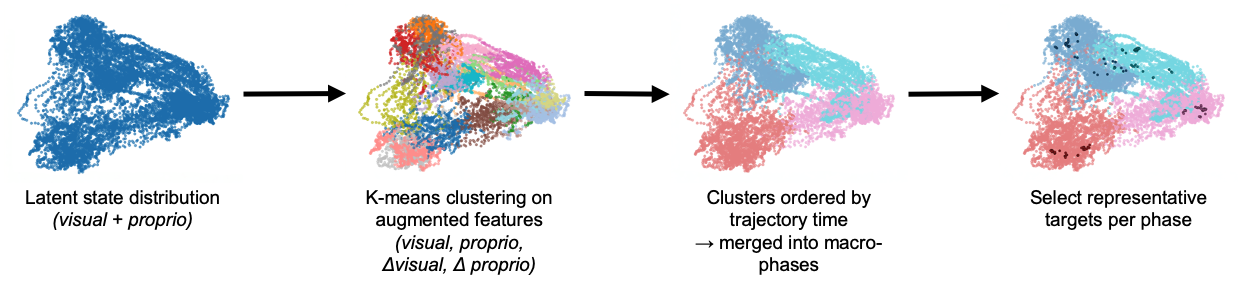}
\caption{\textbf{Phase-aware target construction.} Latents states are clustered using temporally augmented features, ordered by trajectory time, and grouped into macro-phases. Representative centroids from each phase define the target sets used for phase-conditioned guidance.}
\label{fig:target_construction}
\vspace{-0.5cm}
\end{figure}

ReGuide requires guidance targets that are precise enough to correct deviations, but not so restrictive that they collapse valid modes of expert behavior. This is especially important in long-horizon manipulation, where visually similar observations may correspond to different temporal stages, and where each stage can contain multiple valid action modes. A single global expert latent set can therefore produce ambiguous guidance: it may pull a rollout toward the wrong stage of the task, or treat a valid mode as out-of-distribution. We address this by constructing phase-conditioned target sets from the current training data. Figure~\ref{fig:target_construction} gives the qualitative summary. Formal description is given in Algorithm~\ref{alg:target-construction}, which is deferred to the Appendix due to page constraint. 

For each state, we form an augmented feature vector
$[v_t,\ p_t,\ \Delta v_t,\ \Delta p_t]$, where $v_t$ and $p_t$ denote visual and proprioceptive latents, and $\Delta v_t$, $\Delta p_t$ are one-step temporal differences. The temporal differences help disambiguate states that are visually similar but belong to different phases of the task. We cluster these features after dimensionality reduction using a standard Principal Component Analysis (PCA), sort clusters by their mean timestep, and merge adjacent clusters into $N_p$ macro-phases. For each phase $j$, we select $M$ representative cluster centroids as the phase target set $\mathcal{P}_j=\{p_{j,1},\ldots,p_{j,M}\}$. Given a predicted latent $z$, its distance to phase $j$ is defined by a soft minimum over phase targets, $\mathcal{L}(z, \mathcal{P}_j)=
-\tau \log \sum_{m=1}^{M}
\exp\left(-\frac{|z-p_{j,m}|^2}{\tau}\right),$

where $\tau$ controls the sharpness of the minimum. This preserves multimodality within each phase while providing a differentiable guidance objective. We also store the empirical distance distribution
$\mathcal{F}_j={\mathcal{L}(z_t, \mathcal{P}_j): z_t \text{ belongs to phase } j}$.
These per-phase distributions are used in Section~\ref{sec:guidance} to calibrate the guidance, so that the decision to guide depends on the local geometry of the current phase rather than on a global distance threshold.

\vspace{-0.2cm}
\subsection{Phase-Conditioned Guidance (PCG) and Rollout Data Collection}
\label{sec:guidance}
\vspace{-0.2cm}
\begin{wrapfigure}{r}{0.5\textwidth}
\vspace{-0.4cm}
\vspace{-\baselineskip}
\begin{minipage}{0.5\textwidth}
\begin{algorithm}[H]
\caption{Phase-Conditioned Guidance and Rollout Collection.}
\label{alg:phase-rollout}
\begin{algorithmic}[1]
\small
\Require Number of macro-phases $N_p$, Phase targets $\{\mathcal{P}_j\}_{j=1}^{N_p}$, guided denoising steps $K_g$, thresholds $\{(\ell_{\text{low}}^{(j)}, \ell_{\text{high}}^{(j)})\}_{j=1}^{N_p}$.
\State $j \gets 1$\; 
\For{$t = 0$ to $T-1$}
    \State Sample $A_t^{K} \sim \mathcal{N}(0, I)$
    \For{$k = K$ down to $1$}
        \State $\hat{\epsilon} \gets \epsilon_\theta(A_t^k, k)$
        \State $\hat{A}_t^0 \gets$ Eq.~\eqref{eq:clean-action}. $\ell \gets \mathcal{L}(d_\phi(z_t, \hat{A}_t^0), \mathcal{P}_j)$
        \If{$k \leq K_g$ \textbf{and} $\ell_{\text{low}}^{(j)} < \ell < \ell_{\text{high}}^{(j)}$}
            \State Update $\hat{\epsilon}$ via Eqs.~\eqref{eq:mpgd-update}--\eqref{eq:guided-noise}
        \EndIf
        \State $A_t^{k-1} \gets$ Eq.~\eqref{eq:reverse-step} with $\hat{\epsilon}$
    \EndFor
    \State Guidance action $A^g_t \gets A_t^0$.  Execute $A^g_t$
    \State Update $j$ per consecutive rule (Sec.~\ref{sec:guidance})
\EndFor
\State \Return $\mathcal{D}^g=\{(O_t, A^g_t)\}_{t=0}^{T-1}$
\end{algorithmic}
\end{algorithm}
\end{minipage}
\vspace{-0.4cm}
\end{wrapfigure}

Given the phase target sets from Section~\ref{sec:target_construction}, ReGuide guides the diffusion policy during rollout generation. The goal is not only to improve the current episode, but to produce trajectories that can be reused for training. For this,  the guidance  must be applied in a form compatible with the learned dynamics model, and only in regions where the model provides reliable gradients.

Prior diffusion-policy guidance methods typically differentiate the 
guidance objective with respect to the noisy denoising iterate $A_t^k$. 
This is a mismatch for our dynamics model $d_\phi$, which is trained on 
clean action chunks. We therefore adapt the MPGD-style update~\citep{he2023mpgd} 
and apply the guidance gradient through the estimated clean action. At 
denoising step $k$, we first compute $\hat{A}_t^0$ via 
Eq.~\eqref{eq:clean-action}. For the current phase $j$, the dynamics 
model predicts the future latent $d_\phi(z_t,\,\hat{A}_t^0)$, and the 
phase-conditioned guidance objective is the soft-minimum distance from 
this latent to $\mathcal{P}_j$. We then update the estimated clean 
action via Eq.~\eqref{eq:mpgd-update}. Since the diffusion scheduler 
expects a noise prediction, we convert the guided clean-action estimate 
back into the corresponding guided noise via Eq.~\eqref{eq:guided-noise}.

\begin{wrapfigure}{r}{0.57\linewidth}
\vspace{-0.7cm}
{\setlength{\jot}{8pt}
\begin{align}
    \label{eq:clean-action}
    \hat{A}_t^0 &= \frac{A_t^k - \sqrt{1-\bar{\alpha}_k}\,\epsilon_\theta}{\sqrt{\bar{\alpha}_k}}, \\
    \label{eq:mpgd-update}
    \tilde{A}_t^0 \leftarrow \hat{A}_t^0 - \eta &\sqrt{1 - \bar{\alpha}_k}\;
    \nabla_{\hat{A}_t^0}\,
    \mathcal{L}\bigl(d_\phi(z_t,\;\hat{A}_t^0),\, \mathcal{P}_j\bigr), \\
    \label{eq:guided-noise}
    \hat{\epsilon} &= \frac{A_t^k - \sqrt{\bar{\alpha}_k}\,\tilde{A}_t^0}{\sqrt{1-\bar{\alpha}_k}}.
\end{align}}
\vspace{-1em}
\end{wrapfigure}

Guidance is useful only when the rollout has deviated from the demonstrations but remains recoverable. If the predicted future latent is already close to the current phase targets, guidance is unnecessary and can perturb a correct action. If it is too far from the phase distribution, the dynamics model is extrapolating, and its gradient may generate low-quality actions. ReGuide, therefore, uses a two-threshold gate for each phase. Let $\ell=\mathcal{L}\bigl(d_\phi(z_t,\;\hat{A}_t^0),\mathcal{P}_j)$ be the phase-conditioned distance at the current denoising step. From the empirical distance distribution $\mathcal{F}_j$, we define lower and upper thresholds $\ell_{\text{low}}^{(j)}$ and $\ell_{\text{high}}^{(j)}$ as fixed percentiles. Guidance is applied only when $\ell_{\text{low}}^{(j)} < \ell < \ell_{\text{high}}^{(j)}$. Thus, the lower threshold avoids unnecessary intervention near the demonstration manifold, while the upper threshold prevents the method from trusting dynamics gradients in extrapolation regions. Because the thresholds are phase-specific, the gate adapts to the local geometry of each stage of the task.

During execution, ReGuide maintains the current phase index $j$. We advance from phase $j$ to $j+1$ when the predicted latent is closer to $\mathcal{P}_{j+1}$ than to $\mathcal{P}_j$ by a margin $\Delta$ for $K$ consecutive rollout steps, which prevents transient prediction noise from causing premature phase switches. After rollout completion, we retain only successful guided trajectories. Each retained trajectory is stored as
$\{(O_t, A^g_t)\}_{t=0}^{T-1}$, where $A^g_t$ is the guided action chunk at step $t$. The trajectory is added to the guided buffer $\mathcal{D}_i^g$, which is merged into the training set for the next ReGuide iteration: $\mathcal{D}_{i+1} \leftarrow \mathcal{D}_i \cup \mathcal{D}_i^g$.

\vspace{-0.2cm}
\subsection{Iterative Self-Improvement Learning}
\label{sec:continual}
\vspace{-0.2cm}

The rollout procedure in Algorithm~\ref{alg:phase-rollout} converts guidance into Dagger-style  training data: at iteration $i$, we run the current policy with phase-conditioned guidance, keep only successful trajectories, and merge them into a guided buffer $\mathcal{D}_i^g$. The updated policy then collects the next batch of guided rollouts, forming a rollout--collect--train loop that can be run for a single round or repeated across multiple iterations; we reuse the dynamics model trained on the base policy's rollouts across all iterations rather than retraining it. We consider three ways to absorb the guided data into the policy. 

\textbf{ReGuide-FT.}
The first variant fine-tunes the current policy checkpoint on a rehearsal buffer containing both prior training data and newly collected guided rollouts. At iteration $i$, minibatches are sampled from the existing dataset $\mathcal{D}_i$ and the guided buffer $\mathcal{D}_i^g$ with a fixed ratio $\rho$. This update is computationally efficient and preserves the competence of the current policy while exposing it to recoverable states reached during guided execution.

\textbf{ReGuide-FS.}
The second variant trains a fresh diffusion policy from scratch on the augmented dataset $\mathcal{D}_i\cup \mathcal{D}_i^g$. This removes dependence on the previous checkpoint and can yield a different solution when the base policy is a poor initialization, at the cost of a full retraining run.

\textbf{ReGuide-FS$\rightarrow$FT.}
Since ReGuide-FT operates on any starting checkpoint, we can stack the variants: apply ReGuide-FT on top of a ReGuide-FS policy rather than the base. ReGuide-FS provides a stronger initialization than the base, so the FT round refines an already-competent policy; the composition costs both runs but yields ReGuide's best results on Can, Square, and Transport in Table~\ref{tab:main_results_compact}.

\section{Experiments}
\label{sec:experiment}

\textbf{Experimental setup.} We evaluate ReGuide on four Robomimic manipulation tasks~\citep{mandlekar2021robomimic}: Can, Square, Transport, and Tool Hang. For each task, we train a diffusion policy on a small subset of the available demonstration data: $15$ demonstrations for Can, $30$ for Square, $10$ for Transport, and $80$ for Tool Hang. This places the base policies in a low-data regime where behavior cloning is susceptible to covariate shift. We train a latent-space dynamics model using rollouts from the corresponding base policy, and construct phase targets and per-phase distance distributions from the current training data. Each reported success rate is computed from $2{,}500$ rollouts of a fixed trained checkpoint, using $50$ initial seeds and $50$ rollouts per seed. We report the mean success rate $\pm$ standard error of the mean. Additional implementation details are given in Appendix~\ref{sec:setup_details}.

\begin{figure}[t]
\centering
\includegraphics[width=0.95\linewidth]{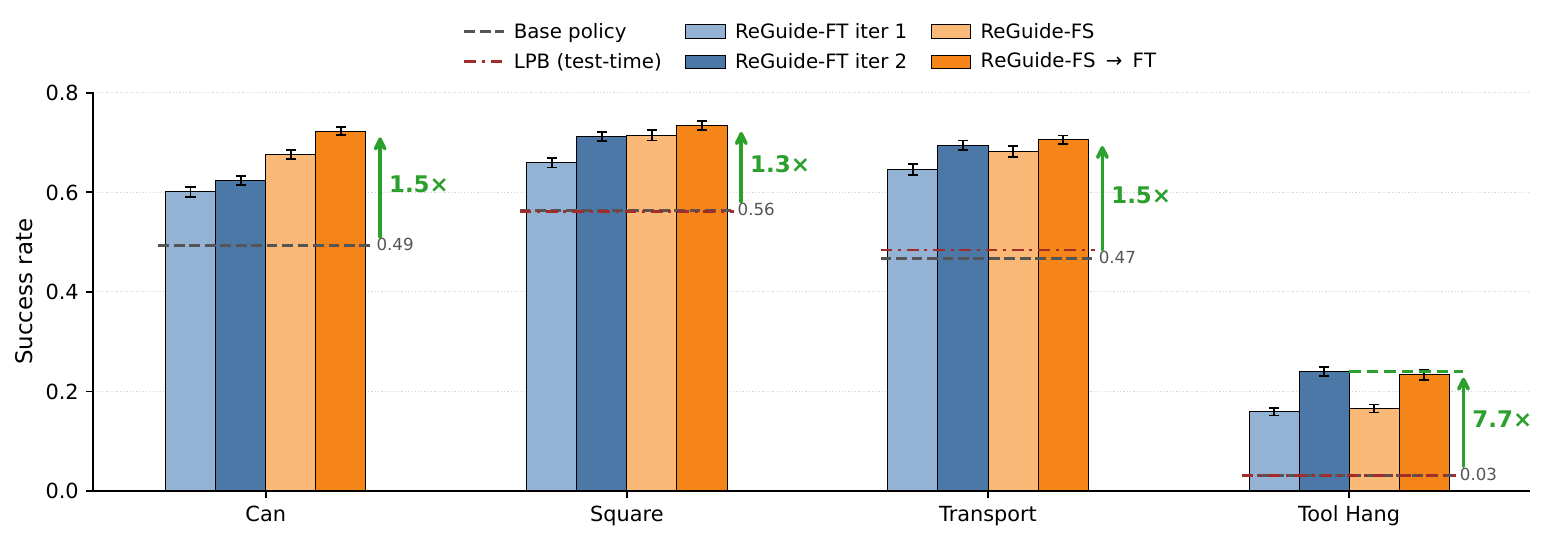}
\caption{\textbf{Main results.} ReGuide improves over the base diffusion policy across all tasks. ReGuide-FT and ReGuide-FS are complementary variants; their composition ReGuide-FS$\rightarrow$FT gives the best result on Can, Square, and Transport, while iterated ReGuide-FT remains slightly stronger on Tool Hang.}
\label{fig:main_results}
\end{figure}

\textbf{Main results.} Figure~\ref{fig:main_results} and Table~\ref{tab:main_results_compact} summarize the main results. All three ReGuide variants improve substantially over the base diffusion policy. \textbf{ReGuide-FT}, fine-tuning from the base checkpoint, lifts success by $1.2$--$1.5\times$ on Can, Square, and Transport, and by $7.7\times$ on Tool Hang after two iterations. \textbf{ReGuide-FS}, retraining from scratch on demonstrations plus guided rollouts, achieves $1.3$--$1.5\times$ on the same three tasks and $5.3\times$ on Tool Hang. Neither single-variant configuration uniformly dominates: ReGuide-FS leads on Can, the two are comparable on Square, and ReGuide-FT's second iteration slightly exceeds ReGuide-FS on both Transport and Tool Hang. \textbf{ReGuide-FS$\rightarrow$FT}, the composition, achieves the best lift on Can, Square, and Transport ($1.5\times$, $1.3\times$, $1.5\times$), confirming that the two variants are complementary: ReGuide-FS moves the policy to a stronger starting checkpoint, while ReGuide-FT efficiently refines it. On Tool Hang, however, the composition ($7.5\times$) lands slightly below ReGuide-FT's second iteration ($7.7\times$). We attribute this to the base policy being weak enough ($3\%$ success) that ReGuide-FS cannot reach a meaningfully stronger starting checkpoint within the limited training budget, so iterating FT extracts more signal from successive guided rollout batches than swapping checkpoints does.

\begin{table}[t]
\vspace{-0.5cm}
\centering
\small
\setlength{\tabcolsep}{4pt}
\caption{Success rates across Robomimic tasks. Our approaches, \textbf{PCG}, \textbf{ReGuide-FT}, \textbf{ReGuide-FS}, and \textbf{ReGuide-FS$\rightarrow$FT}, all outperform the base policy and LPB. The composition ReGuide-FS$\rightarrow$FT achieves the best result on Can, Square, and Transport; iterated ReGuide-FT remains slightly stronger on Tool Hang. For each task, the highest success rate is in \textbf{bold} and the second-best is \underline{underlined}.}
\label{tab:main_results_compact}
\resizebox{\linewidth}{!}{%
\begin{tabular}{lccccccc}
\toprule
& \multicolumn{3}{c}{Test-Time Guidance} & \multicolumn{2}{c}{ReGuide-FT} & & \\
\cmidrule(lr){2-4} \cmidrule(lr){5-6}
Task
& Base Policy & LPB~\citep{sun2025lpb} & PCG
& Iteration 1 & Iteration 2
& ReGuide-FS & ReGuide-FS $\rightarrow$ FT \\
\midrule
Can
& $0.4924$\,\scriptsize{\textcolor{gray}{$\pm$\,0.0101}} & --- & $0.5012$\,\scriptsize{\textcolor{gray}{$\pm$\,0.0100}}
& $0.6008$\,\scriptsize{\textcolor{gray}{$\pm$\,0.0098}} & $0.6236$\,\scriptsize{\textcolor{gray}{$\pm$\,0.0090}}
& $\underline{0.6764}$\,\scriptsize{\textcolor{gray}{$\pm$\,0.0089}} & $\textbf{0.7228}$\,\scriptsize{\textcolor{gray}{$\pm$\,0.0083}} \\

Square
& $0.5632$\,\scriptsize{\textcolor{gray}{$\pm$\,0.0091}} & $0.5608$\,\scriptsize{\textcolor{gray}{$\pm$\,0.0099}} & $0.5836$\,\scriptsize{\textcolor{gray}{$\pm$\,0.0104}}
& $0.6592$\,\scriptsize{\textcolor{gray}{$\pm$\,0.0095}} & $0.712$\,\scriptsize{\textcolor{gray}{$\pm$\,0.0085}}
& $\underline{0.7144}$\,\scriptsize{\textcolor{gray}{$\pm$\,0.0101}} & $\textbf{0.7336}$\,\scriptsize{\textcolor{gray}{$\pm$\,0.0091}} \\

Transport
& $0.4664$\,\scriptsize{\textcolor{gray}{$\pm$\,0.0086}} & $0.4840$\,\scriptsize{\textcolor{gray}{$\pm$\,0.0082}} & $0.5140$\,\scriptsize{\textcolor{gray}{$\pm$\,0.0102}}
& $0.6456$\,\scriptsize{\textcolor{gray}{$\pm$\,0.0110}} & $\underline{0.6948}$\,\scriptsize{\textcolor{gray}{$\pm$\,0.0094}}
& $0.6820$\,\scriptsize{\textcolor{gray}{$\pm$\,0.0114}} & $\textbf{0.7056}$\,\scriptsize{\textcolor{gray}{$\pm$\,0.0086}} \\

Tool Hang
& $0.0312$\,\scriptsize{\textcolor{gray}{$\pm$\,0.0030}} & $0.0312$\,\scriptsize{\textcolor{gray}{$\pm$\,0.0036}} & $0.0384$\,\scriptsize{\textcolor{gray}{$\pm$\,0.0036}}
& $0.1592$\,\scriptsize{\textcolor{gray}{$\pm$\,0.0077}} & $\textbf{0.2404}$\,\scriptsize{\textcolor{gray}{$\pm$\,0.0088}}
& $\underline{0.1656}$\,\scriptsize{\textcolor{gray}{$\pm$\,0.0078}} & $0.2332$\,\scriptsize{\textcolor{gray}{$\pm$\,0.0106}}\\

\bottomrule
\end{tabular}%
}
\vspace{-0.5cm}
\end{table}

\begin{figure}[t]
\centering
\includegraphics[width=\linewidth]{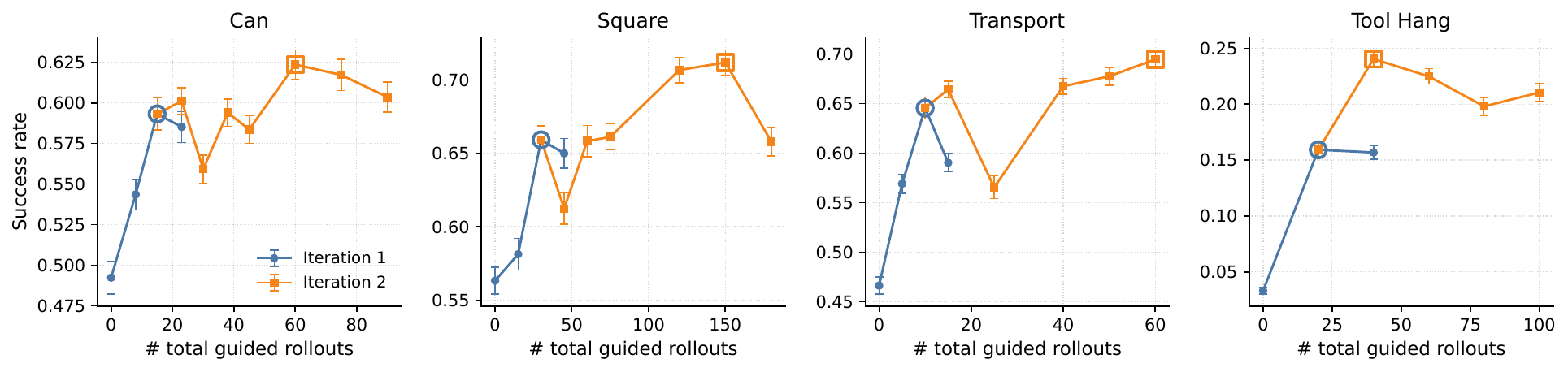}
\caption{\textbf{Iterative self-improvement.} The second iteration of ReGuide-FT improves over the first on all tasks, showing that updated policies can generate useful new guided rollouts. The x-axis shows the cumulative number of guided rollouts collected to update the policy.}
\label{fig:iteration}
\end{figure}

\textbf{Iterative self-improvement.}
Figure~\ref{fig:iteration} shows the success rate as a function of the number of guided rollouts collected for updating the policy for iteration 1 and iteration 2 of ReGuide-FT. Our goal is to evaluate whether the rollout--collect--train loop continues to improve after one round. Starting from the best first-iteration ReGuide-FT policy, we collect a new batch of guided rollouts and perform a second ReGuide-FT iteration. The second iteration outperforms the first on all four tasks: ReGuide-FT's lift over the base grows from $1.2\times$ to $1.3\times$ on Can, $1.2\times$ to $1.3\times$ on Square, $1.4\times$ to $1.5\times$ on Transport, and $5.1\times$ to $7.7\times$ on Tool Hang. The gains are not unbounded: reaching the second-iteration peak requires more total guided rollouts, suggesting diminishing returns as the policy improves and the fixed dynamics model becomes less aligned with the updated policy. 

\textbf{Comparison between LPB and PCG.} We compare our PCG apporach to LPB~\citep{sun2025lpb} in the test-time-only setting (no finetuning) in Table~\ref{tab:main_results_compact}. PCG wins on every task where LPB is evaluated. These test-time gains are much smaller than the full ReGuide gains in Table~\ref{tab:main_results_compact}, which is expected: the main role of PCG is not to improve a single rollout but to generate cleaner data for the subsequent ReGuide-FT and ReGuide-FS updates, where the $1.3$--$7.7\times$ lifts come from. Other guidance methods (DynaGuide~\citep{du2025dynaguide}, PPGuide~\citep{wang2026ppguide}, GPC~\citep{qi2026gpc}) use different guidance objectives or supervision assumptions, while data-augmentation methods such as CCIL~\citep{ke2024ccil} operate outside the inference-time guidance setting; we therefore treat LPB as the primary apples-to-apples comparison.

\paragraph{Ablation studies.}
The appendix isolates the main algorithmic choices in ReGuide:
\begin{enumerate}
\item \textbf{Number of phase targets.} Table~\ref{tab:transport_phase_target_ablation} sweeps the number of targets per phase $M$ on Transport. A single target ($M=1$) does not improve over the base policy and very large $M$ also underperforms; an intermediate $M\!\approx\!50$ is best, supporting the need to preserve multimodality without diluting the guidance signal.

\item \textbf{Two-threshold guidance gate.} Table~\ref{tab:transport_threshold_design} compares gating rules. A lower-only gate (matching prior work) lifts success by $+0.014$ over no guidance, while the full lower-and-upper gate lifts by $+0.044$---the upper threshold accounts for most of the benefit, supporting the three-regime design (guide only when drifted but recoverable).

\item \textbf{Clean-action guidance.} Table~\ref{tab:transport_guidance_target} compares differentiating the guidance objective through the noisy iterate $A_t^k$ versus the estimated clean action $\hat{A}_t^0$. Clean-action guidance wins by $\approx 2$ SE on Transport, consistent with the dynamics model being trained on clean action chunks.

\item \textbf{Guidance vs.\ additional data.} Tables~\ref{tab:guidance_vs_base_compact} and~\ref{tab:can_cl_guidance_vs_base} (with per-task detail in Tables~\ref{tab:can_rollout_count}--\ref{tab:transport_rollout_count}) compare guided rollouts with unguided base-policy rollouts at matched data volume. Guided rollouts outperform unguided rollouts under both ReGuide-FT and ReGuide-FS, showing ReGuide's gains are not merely from adding more rollouts.

\item \textbf{ReGuide-FT vs.\ ReGuide-FS.} Table~\ref{tab:can_cl_vs_scratch} compares the two absorption mechanisms after the composition step on Can. The two are statistically comparable at matched rollout counts, while ReGuide-FT is more compute-efficient from an already-strong checkpoint.

\item \textbf{Per-task rollout-count sweeps.} Tables~\ref{tab:cl_from_scratch_cross_task} and~\ref{tab:cl_iter_pair} give the per-task sweeps behind ReGuide-FS and the two-iteration ReGuide-FT curves in Figure~\ref{fig:iteration}, showing where each task saturates and the rollout-count vs.\ marginal-gain tradeoff.
\end{enumerate}

\section{Conclusion and Limitations}

ReGuide turns test-time guidance from a one-time inference correction into reusable on-policy recovery data. A successful guided rollout captures where a behavior-cloned policy drifts and which actions recover task progress. ReGuide makes this usable for training through Phase-Conditioned Guidance, which localizes targets to task phases, gates guidance to the drifted-but-recoverable regime, and applies gradients through the estimated clean action. The successful rollouts are then absorbed through ReGuide-FT, ReGuide-FS, or their composition. Across Robomimic tasks, ReGuide improves base-policy success by $1.3$--$7.7\times$, outperforms LPB in the test-time-only setting, and continues improving across rollout--collect--train iterations.

\paragraph{Limitations.}
ReGuide has several limitations, which lead to natural extensions and future works. First, the dynamics model is reused across iterations for efficiency, but refreshing or uncertainty-calibrating it may improve later rounds. Second, rollout selection uses trajectory-level success filtering and random sampling; diversity-aware or quality-aware selection could improve data efficiency. Third, phase construction and gating thresholds are calibrated per task, suggesting future work on adaptive phase discovery and threshold updates. Finally, our evaluation uses fixed checkpoints and simulated Robomimic tasks; broader training-seed studies and real-robot validation are important next steps.

\newpage
\bibliography{reference}

\appendix

\section{Extended Related Work}
\label{sec:extended_related_work}

This section expands on the related work discussion in Section~\ref{sec:related_work}, providing additional detail on individual methods and covering directions that did not fit in the main paper.

\paragraph{Test-time guidance for diffusion policies.} DynaGuide~\citep{du2025dynaguide} uses an external dynamics model to compute gradients that bias the denoising process toward a user-specified goal state, with guidance applied to the noisy denoising iterate at each reverse-diffusion step. Latent Policy Barrier (LPB)~\citep{sun2025lpb} adapts the same gradient mechanism to keep rollouts close to the expert distribution, treating expert latent embeddings as an implicit in-distribution barrier rather than as a goal to reach. PPGuide~\citep{wang2026ppguide} replaces the dynamics-gradient signal with a learned performance predictor, steering toward action chunks that score well under the predictor. Generative Predictive Control (GPC)~\citep{qi2026gpc} takes a different approach: rather than gradient-based guidance, it samples multiple action candidates and uses an action-conditioned world model to rank them via lightweight model-based look-ahead. PCG differs from these methods along three axes: (i) it differentiates through the dynamics model with respect to the estimated clean action rather than the noisy iterate, matching the dynamics model's training distribution; (ii) it replaces a single global target set with phase-conditioned target sets that preserve expert multimodality within each phase; and (iii) it gates guidance to the drifted-but-recoverable regime via a two-threshold rule calibrated to per-phase distance distributions.

\paragraph{Data augmentation for imitation learning.} Beyond CCIL discussed in the main paper, several methods expand the training distribution offline using task structure or assumptions about which perturbations preserve correctness. JUICER~\citep{ankile2024juicer} combines expressive architectures with dataset-expansion and simulation-based augmentation for long-horizon assembly. SEIL~\citep{jia2022seil} supplements expert trajectories with simulated transitions and exploits manipulation symmetries to generate equivalent demonstrations. Stable-BC~\citep{mehta2025stablebc} regularizes the policy so that its closed-loop behavior is locally stable around expert states. These methods share with ReGuide the goal of expanding training coverage without additional expert supervision, but rely on assumption-driven synthesis: known symmetries, local dynamics models, or stability conditions. ReGuide instead generates corrective data \emph{on-policy} by rolling out the current policy under guidance and keeping successful trajectories, requiring only a trajectory-level success signal rather than assumptions about which perturbations preserve task correctness.

\paragraph{Phase / sub-task decomposition in manipulation.} Decomposing long-horizon manipulation into temporally ordered phases or sub-skills is a recurring theme in robot learning, instantiated at several different levels of the pipeline. IRIS~\citep{mandlekar2020iris} factors the control problem into a high-level goal planner and a low-level goal-conditioned controller, learning sub-policies from offline demonstrations. Relay Policy Learning~\citep{gupta2020rpl} learns goal-conditioned hierarchical policies from unsegmented demonstrations via a data-relabeling trick and then fine-tunes them with reinforcement learning. PerAct~\citep{shridhar2022peract} sidesteps dense trajectory prediction altogether by predicting discrete keyframe end-effector poses that serve as phase anchors, with a motion planner handling the low-level control between them. MimicGen~\citep{mandlekar2023mimicgen} uses an object-centric subtask decomposition for offline data generation, transforming and stitching demonstration segments to synthesize new trajectories. JUICER~\citep{ankile2024juicer} similarly leverages subtask structure for both architecture design and augmentation in long-horizon assembly. ReGuide uses phases differently from all of these: the diffusion policy itself stays unfactored, the action representation stays continuous, and the demonstration data is not segmented for retargeting. Phases here serve only as an inference-time tool for localizing the ``what counts as in-distribution'' check that gates our guidance, adapting the threshold to local task geometry without imposing a structural decomposition on the policy, the data, or the action space.

\paragraph{Iterative self-improvement and rehearsal-based learning.} Our rollout--collect--train loop is closer in spirit to self-training in supervised learning~\citep{xie2020noisystudent}, where a model labels new data, filters by confidence, and retrains, than to classical continual learning, even though we borrow the rehearsal buffer construction from the latter. Rehearsal-based continual learning methods retain or replay a subset of past data during new updates to mitigate forgetting~\citep{rebuffi2017icarl, buzzega2020dark}; our two-buffer formulation in Section~\ref{sec:continual} reuses this trick, with the buffer-share ratio $\rho$ functioning as a rehearsal fraction. The setting differs, though: continual learning typically targets sequential acquisition of \emph{distinct} tasks, while we iteratively refine a policy on the \emph{same} task using progressively higher-quality self-collected data. The closest neighbors are therefore self-training methods that filter generated data by an external success signal, with ReGuide's distinguishing element being that the data-generation step is itself a guided diffusion rollout rather than a forward pass of the current model.
\section{Algorithm}
\label{sec:algorithm_appendix}

\begin{algorithm}[H]
\caption{Phase-Aware Target Construction}
\label{alg:target-construction}
\begin{algorithmic}[1]
\Require Training data $\mathcal{D}_{\text{training}} = \{(o_t, a_t)\}$, visual encoder $E_v$, proprio encoder $E_p$, number of clusters $k$, number of macro-phases $N_p$, targets per phase $M$, PCA dimension $d$
\Ensure Per-phase phase target sets $\{\mathcal{P}_j\}_{j=1}^{N_p}$ and distance distributions $\{\mathcal{F}_j\}_{j=1}^{N_p}$
\State \textbf{Encode states with temporal augmentation:}
\For{each $(o_t, a_t) \in \mathcal{D}_{\text{training}}$}
    \State $v_t \gets E_v(o_t)$, \; $p_t \gets E_p(o_t)$ \Comment{visual and proprio features}
    \State $\Delta v_t \gets v_t - v_{t-1}$, \; $\Delta p_t \gets p_t - p_{t-1}$ \Comment{zero at $t=0$}
    \State $z_t \gets [v_t,\; p_t,\; \Delta v_t,\; \Delta p_t]$
\EndFor
\State $\{\tilde{z}_t\} \gets \text{PCA}(\{z_t\}, d)$
\State $\{c_t\} \gets \text{KMeans}(\{\tilde{z}_t\}, k)$
\State \textbf{Order clusters temporally:}
\For{each cluster $c \in \{1, \dots, k\}$}
    \State $\bar{t}_c \gets \text{mean}\{t : c_t = c\}$
\EndFor
\State Sort clusters by $\bar{t}_c$ in ascending order
\State \textbf{Merge into macro-phases:} Group the $k$ ordered clusters into $N_p$ contiguous macro-phases $\{\Phi_1, \dots, \Phi_{N_P}\}$
\State \textbf{Select phase targets and build distance distributions:}
\For{$j = 1$ to $N_p$}
    \State $\mathcal{P}_j \gets$ select $M$ representative latents from $\Phi_j$ (cluster centroids within $\Phi_j$)
    \State $\mathcal{F}_j \gets \{\mathcal{L}(z_t, \mathcal{P}_j) : t \in \Phi_j\}$
\EndFor
\State \Return $\{\mathcal{P}_j\}_{j=1}^{N_p}, \{\mathcal{F}_j\}_{j=1}^{N_p}$
\end{algorithmic}
\end{algorithm}

\section{Ablation Studies}
\label{sec:ablations_guidance}

This appendix collects all ablation studies referenced from the main paper:
\begin{itemize}[noitemsep, topsep=0pt, left=1em]
    \item \textbf{PCG design choices} (below): number of targets per phase (Table~\ref{tab:transport_phase_target_ablation}), two-threshold gating rule (Table~\ref{tab:transport_threshold_design}), and guidance target---clean action vs.\ noisy iterate (Table~\ref{tab:transport_guidance_target}). All on Transport.
    \item \textbf{Guidance vs.\ additional data} (Section~\ref{sec:guidance_in_cl}): matched-data comparison of guided vs.\ unguided rollouts under both ReGuide-FT and ReGuide-FS, across Can, Square, and Transport.
    \item \textbf{ReGuide-FT vs.\ ReGuide-FS} (Section~\ref{sec:cl_vs_scratch}): the two variants applied on top of the same checkpoint at matched rollout counts on Can.
    \item \textbf{Per-task rollout-count sweeps}: ReGuide-FS per task (Section~\ref{sec:train_from_scratch}), the composition across tasks, and two-iteration ReGuide-FT.
    \item \textbf{Buffer-share ratio} (Table~\ref{tab:rho_ablation}): sensitivity of ReGuide-FT to $\rho$ on Can.
\end{itemize}

\paragraph{Number of Phase targets} In Section~\ref{sec:target_construction}, we construct multiple targets per phase to respect the multimodal nature of expert manipulation behavior. Table~\ref{tab:transport_phase_target_ablation} evaluates the effect of this choice on Transport. Performance is non-monotone in $M$: $M=1$ (one target per phase) and very large $M$ both underperform; $M=50$ achieves the best success rate.

Intuitively, too few phase targets over-commit the policy to whichever single modes happen to be captured, while too many dilute the gradient signal across loosely-related targets; the middle range balances these effects.

\paragraph{Guidance Threshold} Table~\ref{tab:transport_threshold_design} compares two gating strategies for test-time guidance: applying guidance whenever the soft-minimum distance exceeds a lower threshold (the standard approach in prior work~\citep{sun2025lpb}), versus our two-threshold rule that additionally disables guidance when the distance exceeds an upper threshold. With both thresholds, guidance improves over the no-guidance baseline by $+0.044$. With only the lower threshold, the improvement shrinks to $+0.014$---essentially at baseline. The upper threshold accounts for the majority of guidance's benefit on Transport, consistent with the three-regime view in Section~\ref{sec:guidance}: when the predicted future state is far from the data manifold, the gradient of distance to distant phase targets pulls toward unreliable targets, washing out the gains from cases where guidance is genuinely useful.

\begin{table}[H]
\centering
\small
\caption{Test-time guidance ablations on Transport, evaluated over $950$ rollouts (19 starting seeds $\times$ 50 rollouts each). Base policy (no guidance): $0.4664$\,\scriptsize{\textcolor{gray}{$\pm$\,0.0140}}\normalsize{. Best result in each sub-table in \textbf{bold}.}}
\label{tab:transport_guidance_ablations}
\begin{subtable}{0.42\textwidth}
\centering
\setlength{\tabcolsep}{4pt}
\begin{tabular}{cc}
\toprule
\# Phase targets $M$ & Success Rate \\
\midrule
1   & $0.4663$\,\scriptsize{\textcolor{gray}{$\pm$\,0.0148}} \\
10  & $0.4653$\,\scriptsize{\textcolor{gray}{$\pm$\,0.0164}} \\
50  & $\mathbf{0.5358}$\,\scriptsize{\textcolor{gray}{$\pm$\,0.0169}} \\
100 & $0.5126$\,\scriptsize{\textcolor{gray}{$\pm$\,0.0131}} \\
150 & $0.4758$\,\scriptsize{\textcolor{gray}{$\pm$\,0.0156}} \\
\bottomrule
\end{tabular}
\caption{Number of phase targets $M$ per phase. Other hyperparameters fixed at $N_p = 4$, $k = 40$, $\tau = 0.7$, thresholds at 50/90.}
\label{tab:transport_phase_target_ablation}
\end{subtable}
\hfill
\begin{subtable}{0.55\textwidth}
\centering
\setlength{\tabcolsep}{4pt}
\begin{tabular}{lc}
\toprule
Gating strategy & Success Rate \\
\midrule
No guidance                                              & $0.4664$\,\scriptsize{\textcolor{gray}{$\pm$\,0.0140}} \\
Lower only ($\ell_{\text{low}}{=}50\%$)                     & $0.4800$\,\scriptsize{\textcolor{gray}{$\pm$\,0.0170}} \\
Upper only ($\ell_{\text{high}}{=}80\%$)                    & $0.4947$\,\scriptsize{\textcolor{gray}{$\pm$\,0.0138}} \\
Both ($\ell_{\text{low}}{=}50\%, \ell_{\text{high}}{=}80\%$)   & $\mathbf{0.5105}$\,\scriptsize{\textcolor{gray}{$\pm$\,0.0131}} \\
\bottomrule
\end{tabular}
\caption{Effect of guidance gating thresholds. With $M = 50$, $N_p = 4$.}
\label{tab:transport_threshold_design}
\end{subtable}
\end{table}

\paragraph{Guidance Target: Clean Action vs.\ Noisy Iterate} Section~\ref{sec:guidance} argues that the gradient guidance should be computed with respect to the estimated clean action $\hat{A}_t^0$ (the MPGD-style choice~\citep{he2023mpgd}) rather than the noisy iterate $A_t^k$ (the formulation used by prior diffusion-policy guidance work~\citep{du2025dynaguide, sun2025lpb}), because our dynamics model $d_\phi$ is trained on clean actions and backpropagating through it with noisy inputs violates its training distribution. Table~\ref{tab:transport_guidance_target} reports the comparison on Transport, averaged over $2{,}000$ rollouts (40 starting seeds $\times$ 50 rollouts each) with all other guidance hyperparameters held fixed. Clean-action guidance outperforms noisy-iterate guidance on average ($0.5135$ vs.\ $0.4890$, a gap of $\approx\!2$ SE), with the direction consistent across seeds and consistent with the MPGD argument: matching the dynamics model's input distribution is what makes its gradient signal informative.

\begin{table}[H]
\centering
\small
\setlength{\tabcolsep}{6pt}
\caption{Effect of the guidance target on Transport: differentiating through the dynamics model with respect to the estimated clean action $\hat{A}_t^0$ (ours, MPGD-style) vs.\ the noisy iterate $A_t^k$ (prior diffusion-policy guidance work). Both share the same phase targets, thresholds, and guidance scale; only the gradient target differs. Mean $\pm$ standard error of the mean over $2{,}000$ rollouts (40 starting seeds $\times$ 50 rollouts each).}
\label{tab:transport_guidance_target}
\begin{tabular}{lcc}
\toprule
Guidance target & \# Seeds & Success Rate \\
\midrule
Noisy iterate $A_t^k$~\citep{du2025dynaguide, sun2025lpb} & 40 & $0.4890$\,\scriptsize{\textcolor{gray}{$\pm$\,0.0118}} \\
Clean action $\hat{A}_t^0$ (MPGD-style, ours)~\citep{he2023mpgd} & 40 & $\mathbf{0.5135}$\,\scriptsize{\textcolor{gray}{$\pm$\,0.0115}} \\
\bottomrule
\end{tabular}
\end{table}

\subsection{Guidance vs.\ Additional Rollouts}
\label{sec:guidance_in_cl}
 
A natural concern is that ReGuide's gains come from extra data rather than guidance specifically. We test this at matched rollout count in both absorption modes of Section~\ref{sec:continual}.

\paragraph{ReGuide-FS.} Appendix Tables~\ref{tab:guidance_vs_base_compact} and~\ref{tab:transport_rollout_count} cover Can, Square, and Transport (Tool Hang was not run for this comparison). At every dataset size and on every task, ReGuide-FS trained on demonstrations plus \emph{guided} rollouts outperforms the same procedure on the same number of unguided base-policy rollouts---by $7$--$10$ points on Can and Square and $\approx 7$ points on Transport at the best dataset size.

\paragraph{ReGuide-FT.} Table~\ref{tab:can_cl_guidance_vs_base} reports the comparison on Can under ReGuide-FT: guided rollouts win at every count, with gains of $1.8$--$4.8$ points. Square and Transport versions are in progress.

\begin{table}[H]
\centering
\small
\setlength{\tabcolsep}{6pt}
\caption{ReGuide-FT on Can with guided vs.\ base-policy rollouts at matched rollout counts. Both conditions use $\rho = 0.8$. Guided rollouts win at every count.}
\label{tab:can_cl_guidance_vs_base}
\begin{tabular}{ccc}
\toprule
\# Added Rollouts & Base-policy rollouts & Guided rollouts \\
\midrule
0 & $0.4924$\,\scriptsize{\textcolor{gray}{$\pm$\,0.0101}} & --- \\
8  & $0.5256$\,\scriptsize{\textcolor{gray}{$\pm$\,0.0111}} & $\mathbf{0.5436}$\,\scriptsize{\textcolor{gray}{$\pm$\,0.0096}} \\
15 & $0.5664$\,\scriptsize{\textcolor{gray}{$\pm$\,0.0093}} & $\mathbf{0.5932}$\,\scriptsize{\textcolor{gray}{$\pm$\,0.0098}} \\
23 & $0.5376$\,\scriptsize{\textcolor{gray}{$\pm$\,0.0101}} & $\mathbf{0.5852}$\,\scriptsize{\textcolor{gray}{$\pm$\,0.0095}} \\
\bottomrule
\end{tabular}
\end{table}

Together, the two modes rule out the ``just more data'' explanation: at matched volume, guidance contributes distinct value, presumably by covering the drifted-but-recoverable region that the base policy reaches but cannot exit alone.
 
\subsection{ReGuide-FT vs.\ ReGuide-FS}
\label{sec:cl_vs_scratch}

The two variants of Section~\ref{sec:continual} play different roles: ReGuide-FS crosses the capacity gap between the base policy and a stronger initialization, while ReGuide-FT refines an already-competent policy. When the loop iterates, that role assignment leaves an engineering question: at each new iteration, should we re-spend compute on another ReGuide-FS run over the accumulated data, or just add another ReGuide-FT round?

Table~\ref{tab:can_cl_vs_scratch} compares both options on Can at matched rollout counts. Starting from the same ReGuide-FS checkpoint and a newly collected batch of guided rollouts $\mathcal{D}_2^g$, we either (a)~run ReGuide-FS on $\mathcal{D}_{\text{demo}} \cup \mathcal{D}_1^g \cup \mathcal{D}_2^g$, or (b)~run ReGuide-FT on the merged buffer. The two methods produce comparable final performance: ReGuide-FT slightly ahead at 8 rollouts, ReGuide-FS slightly ahead at 15. The gap in either direction is within one SE.

\begin{table}[H]
\centering
\small
\setlength{\tabcolsep}{6pt}
\caption{ReGuide-FT vs.\ ReGuide-FS on Can at matched rollout counts, both starting from the same ReGuide-FS checkpoint (success rate $0.6764$, row 0) and seeing the same newly collected guided rollouts; ReGuide-FT uses $\rho = 0.8$. Performance is statistically comparable; ReGuide-FT wins on compute.}
\label{tab:can_cl_vs_scratch}
\begin{tabular}{ccc}
\toprule
\# Added Rollouts & ReGuide-FT ($\rho=0.8$) & ReGuide-FS \\
\midrule
0  & $0.6764$\,\scriptsize{\textcolor{gray}{$\pm$\,0.0089}} & --- \\
8  & $\mathbf{0.7208}$\,\scriptsize{\textcolor{gray}{$\pm$\,0.0086}} & $0.6878$\,\scriptsize{\textcolor{gray}{$\pm$\,0.0078}} \\
15 & $0.6916$\,\scriptsize{\textcolor{gray}{$\pm$\,0.0081}}           & $\mathbf{0.7088}$\,\scriptsize{\textcolor{gray}{$\pm$\,0.0087}} \\
\bottomrule
\end{tabular}
\end{table}

Given comparable performance, ReGuide-FT is the better engineering choice for iterations beyond the composition step: it converges in far fewer steps from an already-competent policy, while ReGuide-FS repeats most of the original training run. This is what makes the iterative loop practical---each additional round is a ReGuide-FT update, not a ReGuide-FS run.

\subsection{ReGuide-FS per Task}
\label{sec:train_from_scratch}

Per-task data for ReGuide-FS: guided vs.\ base-policy rollouts at matched dataset sizes.

\begin{table}[H]
\centering
\small
\caption{Success rates comparing guided vs.\ base-policy rollouts at matched dataset sizes under ReGuide-FS (Can and Square). Rollouts are merged with the demonstration data and used to train a fresh policy. Guided rollouts outperform base-policy rollouts at every dataset size. Best result per task in \textbf{bold}.}
\label{tab:guidance_vs_base_compact}
\begin{subtable}{0.48\textwidth}
\centering
\setlength{\tabcolsep}{6pt}
\begin{tabular}{ccc}
\toprule
\# Rollouts & Base-policy & Guided \\
\midrule
0  & $0.4924$\,\scriptsize{\textcolor{gray}{$\pm$\,0.0101}} & --- \\
8  & $0.5086$\,\scriptsize{\textcolor{gray}{$\pm$\,0.0107}} & $0.5324$\,\scriptsize{\textcolor{gray}{$\pm$\,0.0100}} \\
15 & $0.5756$\,\scriptsize{\textcolor{gray}{$\pm$\,0.0094}} & $\mathbf{0.6764}$\,\scriptsize{\textcolor{gray}{$\pm$\,0.0089}} \\
23 & $0.5528$\,\scriptsize{\textcolor{gray}{$\pm$\,0.0093}} & $0.6264$\,\scriptsize{\textcolor{gray}{$\pm$\,0.0094}} \\
\bottomrule
\end{tabular}
\caption{Can.}
\label{tab:can_rollout_count}
\end{subtable}
\hfill
\begin{subtable}{0.48\textwidth}
\centering
\setlength{\tabcolsep}{6pt}
\begin{tabular}{ccc}
\toprule
\# Rollouts & Base-policy & Guided \\
\midrule
0  & $0.5632$\,\scriptsize{\textcolor{gray}{$\pm$\,0.0091}} & --- \\
15 & $0.5628$\,\scriptsize{\textcolor{gray}{$\pm$\,0.0101}} & $0.6908$\,\scriptsize{\textcolor{gray}{$\pm$\,0.0096}} \\
30 & $0.6488$\,\scriptsize{\textcolor{gray}{$\pm$\,0.0083}} & $\mathbf{0.7144}$\,\scriptsize{\textcolor{gray}{$\pm$\,0.0101}} \\
45 & $0.6476$\,\scriptsize{\textcolor{gray}{$\pm$\,0.0094}} & $0.6888$\,\scriptsize{\textcolor{gray}{$\pm$\,0.0075}} \\
\bottomrule
\end{tabular}
\caption{Square.}
\label{tab:square_rollout_count}
\end{subtable}
\end{table}

\begin{table}[H]
\centering
\small
\setlength{\tabcolsep}{6pt}
\caption{Success rates on Transport comparing guided vs.\ base-policy rollouts at matched dataset sizes under ReGuide-FS. Best result in \textbf{bold}.}
\label{tab:transport_rollout_count}
\begin{tabular}{ccc}
\toprule
\# Added Rollouts & Base-policy & Guided \\
\midrule
0  & $0.4664$\,\scriptsize{\textcolor{gray}{$\pm$\,0.0086}} & --- \\
5  & $0.5436$\,\scriptsize{\textcolor{gray}{$\pm$\,0.0114}} & $0.6324$\,\scriptsize{\textcolor{gray}{$\pm$\,0.0104}} \\
10 & $0.5924$\,\scriptsize{\textcolor{gray}{$\pm$\,0.0092}} & $\mathbf{0.6820}$\,\scriptsize{\textcolor{gray}{$\pm$\,0.0114}} \\
15 & $0.6080$\,\scriptsize{\textcolor{gray}{$\pm$\,0.0096}} & $0.6544$\,\scriptsize{\textcolor{gray}{$\pm$\,0.0096}} \\
\bottomrule
\end{tabular}
\end{table}

\subsection{Composition (ReGuide-FT on top of ReGuide-FS) Across Tasks}

Per-task detail for the composition: ReGuide-FT starting from the best ReGuide-FS checkpoint, across all four Robomimic tasks.

\begin{table}[H]
\centering
\small
\setlength{\tabcolsep}{6pt}
\caption{ReGuide-FT applied across tasks starting from the best ReGuide-FS checkpoint (the composition). ``Best Scratch'' column gives the starting ReGuide-FS policy. Best result per task in \textbf{bold}.}
\label{tab:cl_from_scratch_cross_task}
\begin{tabular}{lccc}
\toprule
Task & Best ReGuide-FS & \# Rollouts & Success Rate  \\
\midrule
\multirow{3}{*}{Can}
 & \multirow{3}{*}{$0.6764$\,\scriptsize{\textcolor{gray}{$\pm$\,0.0089}}}
 & 8  & $0.7208$\,\scriptsize{\textcolor{gray}{$\pm$\,0.0086}}  \\
 & & 15 & $0.6916$\,\scriptsize{\textcolor{gray}{$\pm$\,0.0081}}  \\
 & & 23 & $\mathbf{0.7228}$\,\scriptsize{\textcolor{gray}{$\pm$\,0.0083}}  \\
\midrule
\multirow{3}{*}{Square}
 & \multirow{3}{*}{$0.7144$\,\scriptsize{\textcolor{gray}{$\pm$\,0.0101}}}
 & 15 & $0.6528$\,\scriptsize{\textcolor{gray}{$\pm$\,0.0110}}  \\
 & & 30 & $0.6596$\,\scriptsize{\textcolor{gray}{$\pm$\,0.0104}}  \\
 & & 45 & $\mathbf{0.7336}$\,\scriptsize{\textcolor{gray}{$\pm$\,0.0091}}  \\
\midrule
\multirow{3}{*}{Transport}
 & \multirow{3}{*}{$0.682$\,\scriptsize{\textcolor{gray}{$\pm$\,0.0114}}}
 & 5  & $0.5896$\,\scriptsize{\textcolor{gray}{$\pm$\,0.0107}}  \\
 & & 10 & $\mathbf{0.7056}$\,\scriptsize{\textcolor{gray}{$\pm$\,0.0086}}  \\
 & & 15 & $0.6664$\,\scriptsize{\textcolor{gray}{$\pm$\,0.0106}}  \\
\midrule
\multirow{4}{*}{Tool Hang}
 & \multirow{4}{*}{$0.1656$\,\scriptsize{\textcolor{gray}{$\pm$\,0.0078}}}
 & 20 & $\mathbf{0.2332}$\,\scriptsize{\textcolor{gray}{$\pm$\,0.0106}} \\
 & & 40 & $0.1776$\,\scriptsize{\textcolor{gray}{$\pm$\,0.0071}} \\
 & & 60 & $0.1976$\,\scriptsize{\textcolor{gray}{$\pm$\,0.0075}} \\
 & & 80 & $0.1836$\,\scriptsize{\textcolor{gray}{$\pm$\,0.0076}} \\
\bottomrule
\end{tabular}
\end{table}

\subsection{Iterating ReGuide-FT from the Base Policy}

Per-task detail across all four tasks for both ReGuide-FT iterations. This is the data underlying Figure~\ref{fig:iteration} and the iteration discussion in the main paper.

\begin{table}[H]
\centering
\small
\caption{Iterative ReGuide-FT from the base policy across tasks: (\textit{top}) Iteration~1, starting from the base diffusion policy; (\textit{bottom}) Iteration~2, starting from the best policy of Iteration~1. \textbf{\# Total Guided Rollouts} counts the cumulative guided rollouts in the training set across both iterations; Iteration~2 rows include the Iteration~1 budget that produced its starting policy (Can: 15, Square: 30, Transport: 10, Tool Hang: 20). Best result per task in \textbf{bold}.}
\label{tab:cl_iter_pair}
\begin{subtable}{\linewidth}
\centering
\setlength{\tabcolsep}{6pt}
\begin{tabular}{lccc}
\toprule
Task & Base & \# Total Guided Rollouts & Success Rate  \\
\midrule
\multirow{3}{*}{Can}
 & \multirow{3}{*}{$0.4924$\,\scriptsize{\textcolor{gray}{$\pm$\,0.0101}}}
 & 8  & $0.5436$\,\scriptsize{\textcolor{gray}{$\pm$\,0.0096}}  \\
 & & 15 & $\mathbf{0.5932}$\,\scriptsize{\textcolor{gray}{$\pm$\,0.0098}}  \\
 & & 23 & $0.5852$\,\scriptsize{\textcolor{gray}{$\pm$\,0.0095}}  \\
\midrule
\multirow{3}{*}{Square}
 & \multirow{3}{*}{$0.5632$\,\scriptsize{\textcolor{gray}{$\pm$\,0.0091}}}
 & 15 & $0.5812$\,\scriptsize{\textcolor{gray}{$\pm$\,0.0107}}  \\
 & & 30 & $\mathbf{0.6592}$\,\scriptsize{\textcolor{gray}{$\pm$\,0.0095}}  \\
 & & 45 & $0.6500$\,\scriptsize{\textcolor{gray}{$\pm$\,0.0101}}  \\
\midrule
\multirow{3}{*}{Transport}
 & \multirow{3}{*}{$0.4664$\,\scriptsize{\textcolor{gray}{$\pm$\,0.0086}}}
 & 5  & $0.5692$\,\scriptsize{\textcolor{gray}{$\pm$\,0.0096}}  \\
 & & 10 & $\mathbf{0.6456}$\,\scriptsize{\textcolor{gray}{$\pm$\,0.0110}}  \\
 & & 15 & $0.5904$\,\scriptsize{\textcolor{gray}{$\pm$\,0.0092}}  \\
\midrule
\multirow{2}{*}{Tool Hang}
 & \multirow{2}{*}{$0.0332$\,\scriptsize{\textcolor{gray}{$\pm$\,0.003}}}
 &  20 & $\mathbf{0.1592}$\,\scriptsize{\textcolor{gray}{$\pm$\,0.007}}  \\
 & & 40 & $0.1568$\,\scriptsize{\textcolor{gray}{$\pm$\,0.006}}  \\
\bottomrule
\end{tabular}
\caption{Iteration~1 (from base policy).}
\label{tab:cl_from_base_cross_task}
\end{subtable}

\vspace{1em}

\begin{subtable}{\linewidth}
\centering
\setlength{\tabcolsep}{6pt}
\begin{tabular}{lccc}
\toprule
Task & Iter.\ 1 Best & \# Total Guided Rollouts & Success Rate \\
\midrule
\multirow{7}{*}{Can}
 & \multirow{7}{*}{$0.5932$\,\scriptsize{\textcolor{gray}{$\pm$\,0.0098}}}
 & 23  & $0.6012$\,\scriptsize{\textcolor{gray}{$\pm$\,0.0082}} \\
 & & 30  & $0.5592$\,\scriptsize{\textcolor{gray}{$\pm$\,0.0087}} \\
 & & 38  & $0.5940$\,\scriptsize{\textcolor{gray}{$\pm$\,0.0084}} \\
 & & 45  & $0.5836$\,\scriptsize{\textcolor{gray}{$\pm$\,0.0086}} \\
 & & 60  & $\mathbf{0.6236}$\,\scriptsize{\textcolor{gray}{$\pm$\,0.0090}} \\
 & & 75  & $0.6172$\,\scriptsize{\textcolor{gray}{$\pm$\,0.0097}} \\
 & & 90  & $0.6036$\,\scriptsize{\textcolor{gray}{$\pm$\,0.0092}} \\
\midrule
\multirow{6}{*}{Square}
 & \multirow{6}{*}{$0.6592$\,\scriptsize{\textcolor{gray}{$\pm$\,0.0095}}}
 & 45  & $0.6124$\,\scriptsize{\textcolor{gray}{$\pm$\,0.0107}} \\
 & & 60  & $0.6584$\,\scriptsize{\textcolor{gray}{$\pm$\,0.0107}} \\
 & & 75  & $0.6612$\,\scriptsize{\textcolor{gray}{$\pm$\,0.0088}} \\
 & & 120 & $0.7068$\,\scriptsize{\textcolor{gray}{$\pm$\,0.0088}} \\
 & & 150 & $\mathbf{0.7120}$\,\scriptsize{\textcolor{gray}{$\pm$\,0.0085}} \\
 & & 180 & $0.6580$\,\scriptsize{\textcolor{gray}{$\pm$\,0.0098}} \\
\midrule
\multirow{5}{*}{Transport}
 & \multirow{5}{*}{$0.6456$\,\scriptsize{\textcolor{gray}{$\pm$\,0.0110}}}
 & 15  & $0.6644$\,\scriptsize{\textcolor{gray}{$\pm$\,0.0082}} \\
 & & 25  & $0.5656$\,\scriptsize{\textcolor{gray}{$\pm$\,0.0115}} \\
 & & 40  & $0.6676$\,\scriptsize{\textcolor{gray}{$\pm$\,0.0081}} \\
 & & 50  & $0.6776$\,\scriptsize{\textcolor{gray}{$\pm$\,0.0089}} \\
 & & 60  & $\mathbf{0.6948}$\,\scriptsize{\textcolor{gray}{$\pm$\,0.0094}} \\
\midrule
\multirow{4}{*}{Tool Hang}
 & \multirow{4}{*}{$0.1592$\,\scriptsize{\textcolor{gray}{$\pm$\,0.007}}}
 & 40  & $\mathbf{0.2404}$\,\scriptsize{\textcolor{gray}{$\pm$\,0.008}} \\
 & & 60  & $0.2248$\,\scriptsize{\textcolor{gray}{$\pm$\,0.007}} \\
 & & 80  & $0.1980$\,\scriptsize{\textcolor{gray}{$\pm$\,0.008}} \\
 & & 100 & $0.2104$\,\scriptsize{\textcolor{gray}{$\pm$\,0.008}} \\
\bottomrule
\end{tabular}
\caption{Iteration~2 (from best of Iteration~1).}
\label{tab:cl_iter2_cross_task}
\end{subtable}
\end{table}

\subsection{Buffer-Share Ratio Ablation}

Ablation over the buffer-share ratio $\rho$ used in ReGuide-FT (Section~\ref{sec:continual}).

\begin{table}[H]
\centering
\small
\setlength{\tabcolsep}{6pt}
\caption{Ablation over buffer-share ratio $\rho$ on Can with 15 added guided rollouts during continual learning, evaluated over $950$ rollouts (19 starting seeds $\times$ 50 rollouts each). Share denotes the percentage of each minibatch drawn from the newly collected guided rollouts.}
\label{tab:rho_ablation}
\begin{tabular}{cc}
\toprule
$\rho$ & Success Rate \\
\midrule
0.6 & $0.5800$\,\scriptsize{\textcolor{gray}{$\pm$\,0.0177}} \\
0.7  & $\mathbf{0.6008}$\,\scriptsize{\textcolor{gray}{$\pm$\,0.0159}} \\
0.8  & $0.5932$\,\scriptsize{\textcolor{gray}{$\pm$\,0.0159}} \\
0.9  & $0.5680$\,\scriptsize{\textcolor{gray}{$\pm$\,0.0146}} \\
1 & $0.5608$\,\scriptsize{\textcolor{gray}{$\pm$\,0.0178}} \\
\bottomrule
\end{tabular}
\end{table}


\section{Setup Details}
\label{sec:setup_details}

This section collects per-task implementation details deferred from the Setup paragraph in Section~\ref{sec:experiment}.

\paragraph{Rollout-count sweeps.} The number of guided rollouts added per iteration is set as a small multiple of the demonstration count for each task: approximately $0.5\times$, $1\times$, and $1.5\times$ in Iteration~1 for Can, Square, and Transport, with higher multiples in Iteration~2 to extend the iteration-2 sweep (see Table~\ref{tab:cl_iter_pair}). Tool Hang uses smaller relative multipliers ($0.25\times$ and $0.5\times$) to keep retraining cost manageable on the largest base policy.

\paragraph{Buffer-share ratio.} For both ReGuide-FT and the FT step of the composition we use a merged buffer with $\rho \in [0.7, 0.8]$, with $\rho$ chosen best-per-task and the newly collected guided rollouts weighted more heavily in each minibatch. The ratio is a stability/plasticity knob: a higher proportion of original demonstration data preserves learned behavior and keeps the update stable but limits how much new behavior the policy can absorb, while a higher proportion of new rollouts accelerates learning of new behavior at the risk of distributional shift and forgetting. The choice is mild within our chosen range---the ablation in Table~\ref{tab:rho_ablation} shows performance is roughly flat on Can.

\paragraph{Supervision assumption.} The rollout-collection step filters for successful trajectories, so the pipeline assumes access to a binary task-success signal at trajectory completion. This is the same assumption made implicitly by any self-improvement scheme that selects among self-collected data, and is strictly weaker than DAgger-style per-state expert relabeling: success-filtering needs one bit per trajectory, not an action label at every visited state. In Robomimic this signal is provided by the simulator; in deployment it would come from a task-specific success classifier, a sparse reward, or human verification at episode end.





\end{document}